# A self-driving lab for solution-processed electrochromic thin films

Selma Dahms[1,†], Luca Torresi[2,3,†], Shahbaz Tareq Bandesha[1], Jan Hansmann[4], Holger Röhm[5], Alexander Colsmann[5], Marco Schott[1,*], and Pascal Friederich[2,3,*]

[1] Fraunhofer Institute for Silicate Research ISC, Neunerplatz 2, 97082 Würzburg, Germany

[2] Institute of Theoretical Informatics, Karlsruhe Institute of Technology, Kaiserstr. 12, 76131 Karlsruhe, Germany

[3] Institute of Nanotechnology, Karlsruhe Institute of Technology, Kaiserstr. 12, 76131 Karlsruhe, Germany

[4] Technical University of Applied Sciences Würzburg-Schweinfurt, Ignaz-Schön-Straße 11, 97421 Schweinfurt, Germany

[5] Material Research Center for Energy Systems, Karlsruhe Institute of Technology, Kaiserstr. 12, 76131 Karlsruhe, Germany

[†] Contributed equally

[*] Corresponding authors: marco.schott@isc.fraunhofer.de and pascal.friederich@kit.edu

Funding: Federal Ministry of Education and Research (BMBF) under Grant No. 01DM21002 (FLAIM). The state of Baden-Württemberg through bwHPC.

## Abstract

Solution-processed electrochromic materials offer high potential for energy-efficient smart windows and displays. Their performance varies with material choice and processing conditions. Electrochromic thin film electrodes require a smooth, defect-free coating for optimal contrast between bleached and colored states. The complexity of optimizing the spin-coated electrochromic thin layer poses challenges for rapid development. This study demonstrates the use of self-driving laboratories to accelerate the development of electrochromic coatings by coupling automation with machine learning. Our system combines automated data acquisition, image processing, spectral analysis, and Bayesian optimization to explore processing parameters efficiently. This approach not only increases throughput but also enables a pointed search for optimal processing parameters. The approach can be applied to various solution-processed materials, highlighting the potential of self-driving labs in enhancing materials discovery and process optimization.

# 1. Introduction

Electrochromic devices have attracted significant attention due to their versatile applications in energy-efficient smart windows and adaptive displays.[1–7] These devices offer dynamic control over light transmittance and color, making them valuable in various sectors. Solution-processed electrochromic (EC) materials present advantages such as scalability and cost-effectiveness, facilitating the production of large-area devices.[8, 9]

However, the development of EC devices (full cells) involves complex, multi-objective optimization of the optical and electrochromic properties of the thin-film electrodes, including color, high contrast ratio, rapid switching speed, cycling stability, and lifetime.[1, 5, 10] Achieving these objectives requires thorough material selection and process optimization to minimize defects and improve performance. The optimization of electrochromic thin films involves the interplay of materials, compositions, and processing parameters. Traditional experimental workflows often rely on manual fabrication and trial-and-error approaches, which are time-consuming and may not efficiently explore the vast parameter space. This underscores the need for systematic and accelerated methodologies to identify optimal processing conditions.

Self-driving laboratories (SDLs) have emerged as a promising approach to address these challenges.[11–16] By integrating automation with closed-loop optimization and machine learning (ML)-driven decision-making, SDLs can enhance reproducibility and efficiency in materials research.[17, 18] In the context of EC materials, SDLs offer the potential to rapidly optimize and discover novel combinations of processing parameters, thereby accelerating development cycles.

We introduce a modular solution-processing platform designed for the automated fabrication of ECDs. It integrates conventional image processing techniques for defect detection, spectral data analysis to assess optical properties, computer vision for automated data analysis, and Bayesian optimization (BO) for data-driven decision-making. We demonstrate high sample reproducibility at high experimental throughput, enabling efficient exploration of the processing parameter space to identify Pareto-optimal samples with both high optical density and low defect density.

While the main focus of this study is on electrochromic materials, the modularity of the platform enables broader applicability to organic electronics and other solution-processed material systems, which we demonstrate as a proof of concept by processing organic light-emitting diode (OLED) thin films. This work aims to contribute to the advancement of SDLs in materials science by demonstrating the use of complex, high-dimensional data modalities, such as spectra and images, to automatically assess sample quality, while rigorously tracking experimental uncertainties and external conditions, thereby enabling the rapid and systematic development of optoelectronic thin films, such as EC and OLED materials.

## 2. Approach

We use an iterative approach that couples experimental thin film processing with thin film characterization (Figure 1A), automated data analysis, and decision making based on BO. [19] This workflow is implemented within a closed-loop framework (Figure 2) where, in each iteration, a set number of samples is prepared, characterized, and used to train surrogate machine learning models for the evaluation and adaptive selection of the most informative experiments, which are then performed in the next iteration.

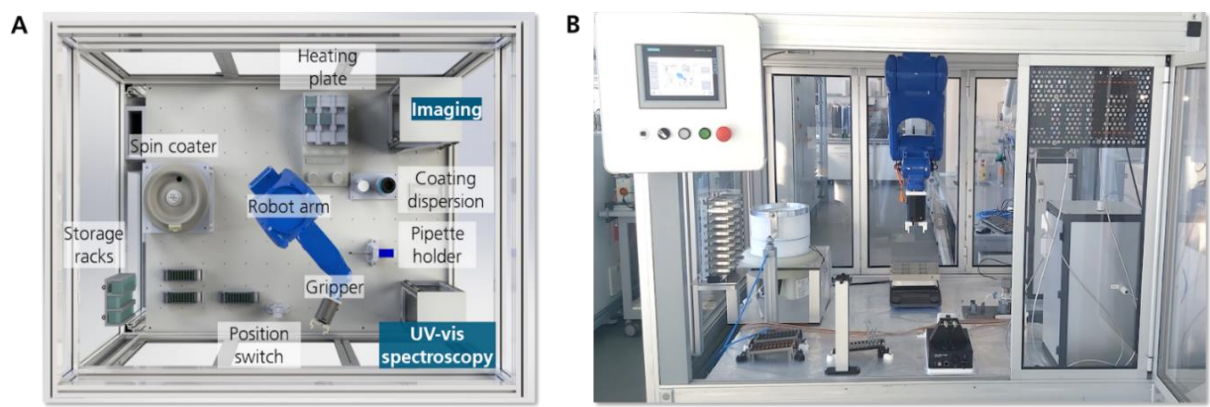

**Figure 1.** Automated platform for the production of (e.g. electrochromic or OLED) thin film materials. (A) Top view of the automated processing and characterization platform, including preparation, storage, processing, and characterization stations. (B) Photograph of the platform, including the central robot arm that transports the sample between different stations and the human-machine interface.

Pareto optimization is employed in this work to maximize optical density while simultaneously minimizing defect density, as improving one objective typically compromises the other. As parameters for the EC layer optimization, we selected the concentration of the coating dispersion (Prussian blue, PB), spin acceleration, spin speed, and spin time. The application volume, annealing time, and temperature were not varied. UV-vis absorbance spectrometry and photographic imaging provide the information required to analyze and optimize the target objectives.

Before coating, the glass substrates with conductive FTO layers were cleaned and treated with an oxygen plasma. PB dispersions were used at various concentrations for spin coating with adjustable spin speeds and spin accelerations on an automated spin coater. After annealing, UV-vis absorbance spectrometry, as well as imaging with either white or black backgrounds, were performed.[20] All stations were controlled by an industrial programmable logic controller and operated in clean room conditions (Figure 1B).

Using classical computer vision techniques, we extracted the relative defect area from the sample images as a proxy for defect density (see Methods: Data Analysis for details). BO was then employed to identify optimal combinations of process parameters that minimize defect density, evaluated separately for images with white or black backgrounds beneath the sample, while maximizing the optical density derived from the absorbance spectra. We employed multi-objective BO to explore the Pareto front of our objectives, that is, the set of solutions balancing low defect density and high optical density. By independently modeling the dependence of each objective on the process parameters, we iteratively selected batches of candidate samples to maximize the Pareto front (for details, see Methods: Bayesian Optimization).

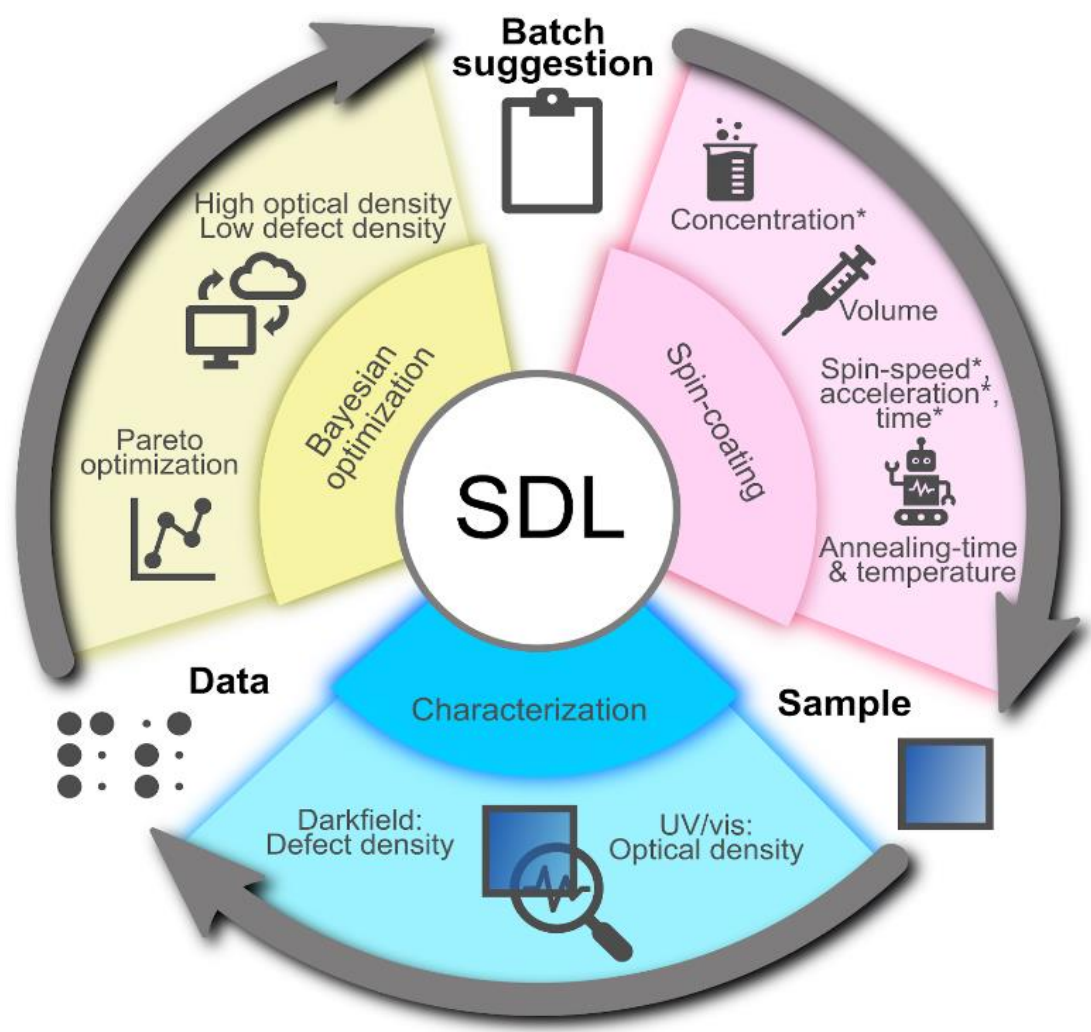

**Figure 2.** Illustration of the optimization loop of Prussian blue thin films by AI and ML methods. An iterative approach was used to find the optimal trade-off between high optical density and low defect density. The optimized parameter set consisted of the parameters highlighted by * in the spin coating section.

## 3. Results and discussion

To maximize the optical density of the PB thin films while minimizing defect density, we conducted a campaign of 10 iterations, each comprising 10 sets of different process parameters. For each set, two samples were produced and evaluated to assess in-batch variability of the experimental process, resulting in a total of 200 samples.

## 3.1 Automated high-throughput thin film processing and characterization

The automated processing and characterization platform under ISO 8 clean room conditions facilitates PB thin film deposition and annealing processes on FTO glass substrates. Using this platform, we fabricated samples with defect densities ranging from 0.042% to 1.47% of the total area and optical densities between 0.15 and 1.55. By performing each experiment twice with identical processing conditions, we estimated the reproducibility of the process by calculating the deviation of each measurement within sample pairs from their midpoint. For 90% of the dataset, defect density deviations ranged from 0.003% to 0.216% (median: 0.043%), and optical density deviations ranged from 0.002 to 0.138 (median: 0.013).

The distributions of defect densities and optical densities across all samples are presented in Figure S1. In all experiments, ambient temperature and air humidity were recorded and taken into account in post-campaign data analysis. We investigated whether the midpoint deviation could be predicted based on the process parameters, including temperature and air humidity; however, no predictive relationship was supported by the data.

The campaign primarily focused on lowering defect density and enhancing optical density; however, the absorption spectrometry provided additional data, including transmittance spectra, visible light transmittance ($\tau v$), and CIE-Lab color coordinates ($L^*, a^*, b^*$), all of which could potentially serve as optimization objectives. These outcomes can be automatically extracted and controlled through the processing parameters used here, as well as further parameters such as ink composition or post-processing. Detailed descriptions of the thin film processing and characterization hardware are available in the Methods section.

## 3.2 Data analysis and decision-making

This study analyzed two complementary types of data: image data to determine defect density and spectral data to determine optical density. Classical computer vision techniques, primarily involving thresholding and filtering operations, were employed to estimate defect density from images. Compared to deep learning, this approach eliminated the need for curated datasets and manual labeling while ensuring high explainability and controllability. However, the reliance on a fixed global grayscale threshold for defect identification makes the method sensitive to variations in ambient light conditions and sample positioning. These issues were mitigated by the use of an automated platform, which ensures repeatability in illumination and sample positioning, and by pre-processing steps, such as image cropping, to further reduce positional variability. Finally, the defect density was estimated using the defect area fraction as a proxy. While this approach cannot differentiate between different sizes of defects, it proved to be adequate for capturing overall sample quality trends and providing a reliable measure for optimization.
The defect density, independently assessed from images using white or black backgrounds, and the optical density, defined at the maximum absorbance of the PB layer in the visible wavelength range, were used as objectives in a multi-objective BO framework. The intervalence charge transfer (IVCT) with a maximum at approximately 710-720 nm is responsible for the characteristic blue color of the mixed-valence compound PB. The acquisition function selected for our Bayesian optimizer is designed to expand the trade-off surface between samples with high optical density and low defect density. The use of Gaussian processes as surrogate models for the mapping between processing parameters and objectives allows capturing nonlinear dependencies among multiple processing parameters. This approach is fast and efficient, enabling the suggestion of optimal parameter sets in batches. The entire data analysis and decision-making process is fully autonomous after initial tuning of a few hyperparameters, such as the reference point defining the minimal acceptable performance conditions on the Pareto front.

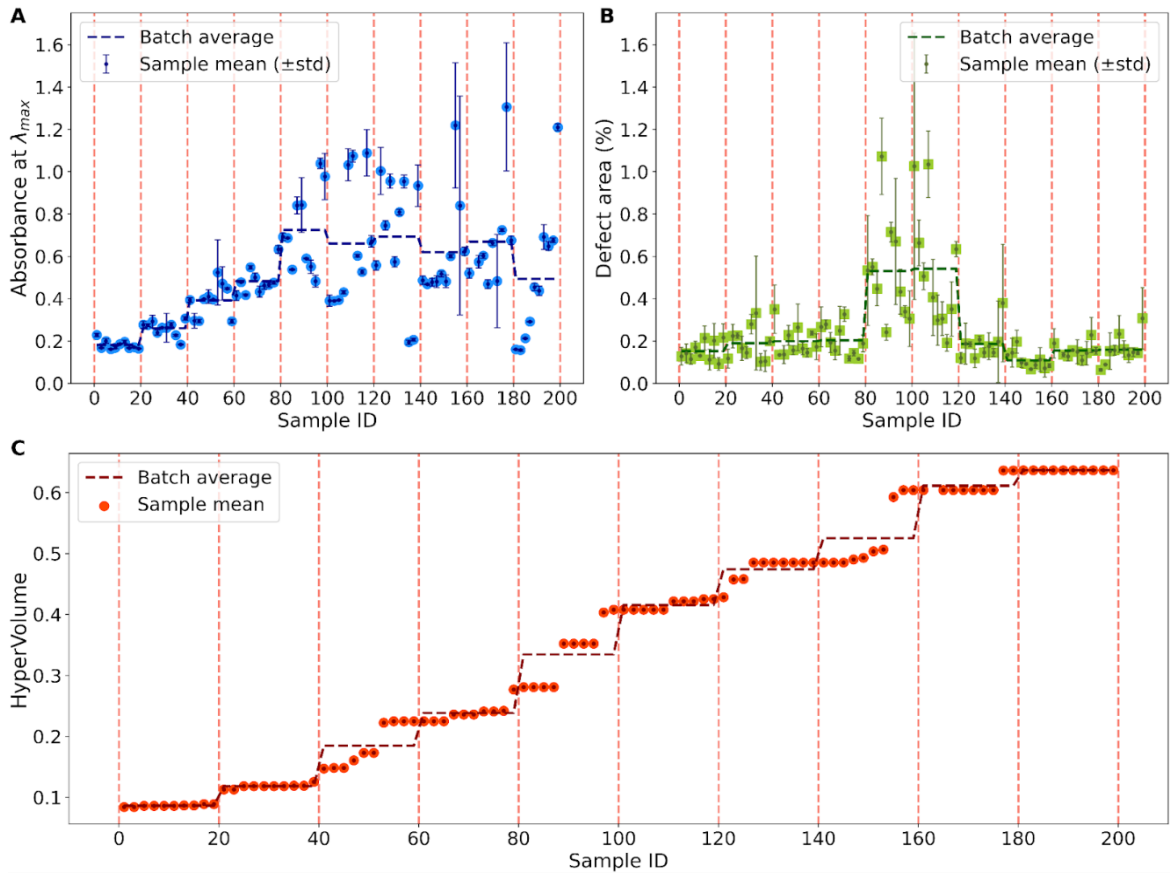

**Figure 3.** Development of (A) absorbance and (B) relative defect area as a function of the sample ID in the Bayesian optimization campaign. Red dashed lines indicated the separation between batches. In each batch, 10 different parameter sets with 2 repetitions each were performed. Error bars indicate the difference in measurements of the two repetitions. Horizontal dashed lines indicate the batch averages. (C) Hypervolume evolution during the multi-objective Bayesian optimization campaign. The hypervolume is defined as the volume spanned in the multi-dimensional objective space relative to a reference point of minimal requirements.

### 3.3 Autonomous exploration of processing parameter space

Figure 3 presents the evolution of key performance indicators across the experimental campaign. Figure 3C illustrates the hypervolume of the Pareto front as a function of the sample

ID, with red dots representing the sample mean and the dashed line indicating the batch average. The hypervolume, a scalar quality indicator, quantifies the volume of the objective space dominated by the set of solutions and provides a measure of the optimization progress. Figure 3A and B display the absorbance at λmax (optical density) and the defect area (summed for bright and dark background images), respectively, for each pair of samples produced with identical process parameters. The dashed lines in panels A and B represent the average per batch. Initially, the first batch of randomly selected process parameters yielded samples with relatively low optical density (Figure 3A). Subsequently, the optimizer successfully increased the optical density of the produced samples in later batches. This increase in optical density, however, was concurrently associated with a rise in defect density (Figure 3B). Towards the latter part of the campaign, the model identified effective combinations of process parameters that led to the fabrication of samples exhibiting both high optical density and low defect density. We also observed that the final batch did not further enhance the hypervolume, suggesting a plateau in the optimization progress (Figure 3C).

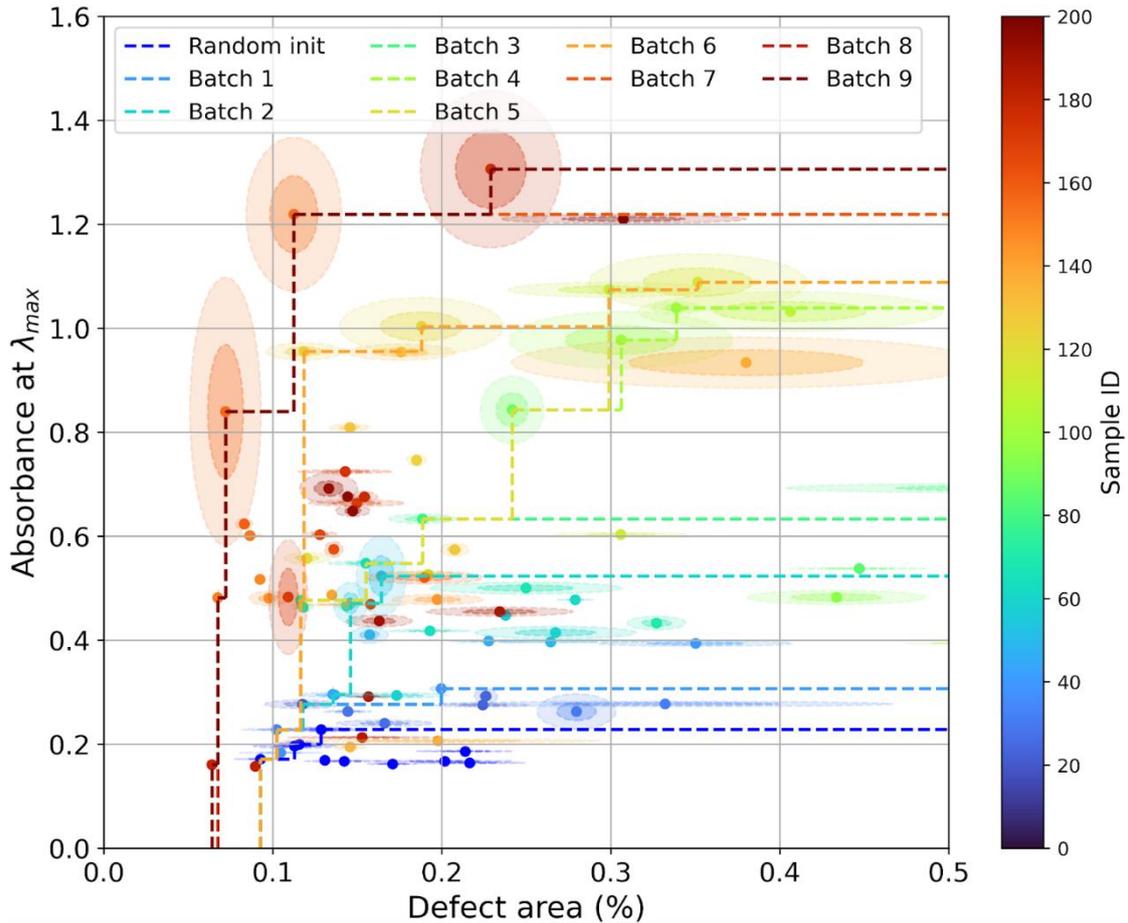

**Figure 4.** Development of the Pareto front of relative defect area and absorbance as a function of sample ID over the entire Bayesian optimization campaign. One dashed line per batch of 20 samples with 10 different parameter sets is shown. The experimental deviation of repetitions with identical parameters is indicated by the shaded ellipses (radii of one and two standard deviations).

Figure 4 offers a more direct visualization of the Pareto front expansion for each batch. The dashed lines delineate the Pareto front, with different colors indicating the specific batch. Each point on the plot represents the mean value for the pair of samples produced with the same processing parameter set, and the surrounding ellipsoids illustrate the relative standard deviation of the measurements. The color map of the points corresponds to the sample ID. Although lower defect densities are generally linked to lower optical densities, the optimal samples discovered in the later iterations demonstrate both high optical densities and low defect densities. Notably, these superior samples also exhibited higher variability in their metrics, as indicated by the larger ellipsoids surrounding their respective data points.

## 3.4 Electrochromic effect of the optimized thin films

The electrochromic switching behavior of the optimized PB thin films is presented by comparing two samples (at the front of the Pareto line) with different optical densities and similar low defect density. These samples were spin coated from an aqueous PB dispersion with a concentration c = 4 wt%.

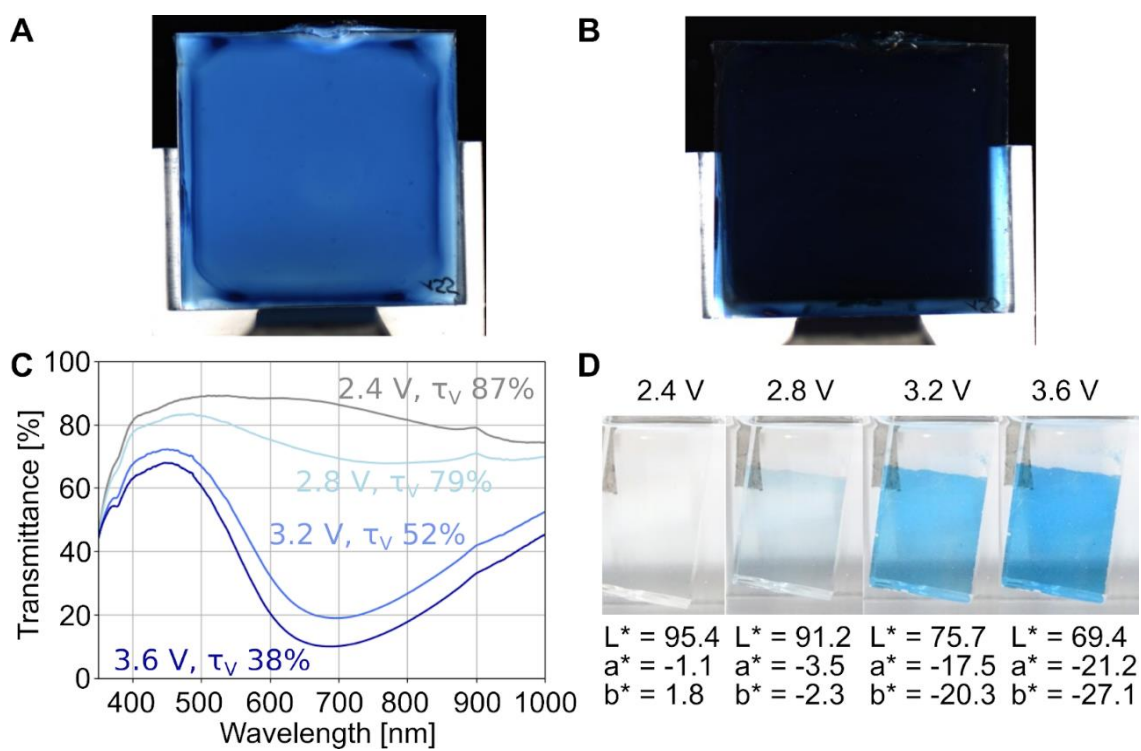

**Figure 5.** Photo of the PB thin film on FTO glass (optical density 1.5, defect density 0.15) against (A) a white background and (B) a black background. (C) Gradual transmittance changes upon application of a potential from 2.4 V (PW) to 3.6 V vs. Li/Li$^+$ (PB) with two intermediate states at 2.8 V and 3.2 V vs. Li/Li$^+$ in 1 M LiClO$_4$/PC. (D) Photos of the PB thin film at 2.4, 2.8, 3.2, and 3.6 V vs. Li/Li$^+$ and the respective L*a*b* color coordinates.

The PB layer with higher optical density (1.5) was spin coated at a spin speed of 500 rpm for 10 s with a spin acceleration of 3000 rpm/s (Figure 5). The PB layer with lower optical density (0.6) was spin coated at a spin speed of 500 rpm for 10 s with a spin acceleration of 1000 rpm/s (Figure 6).

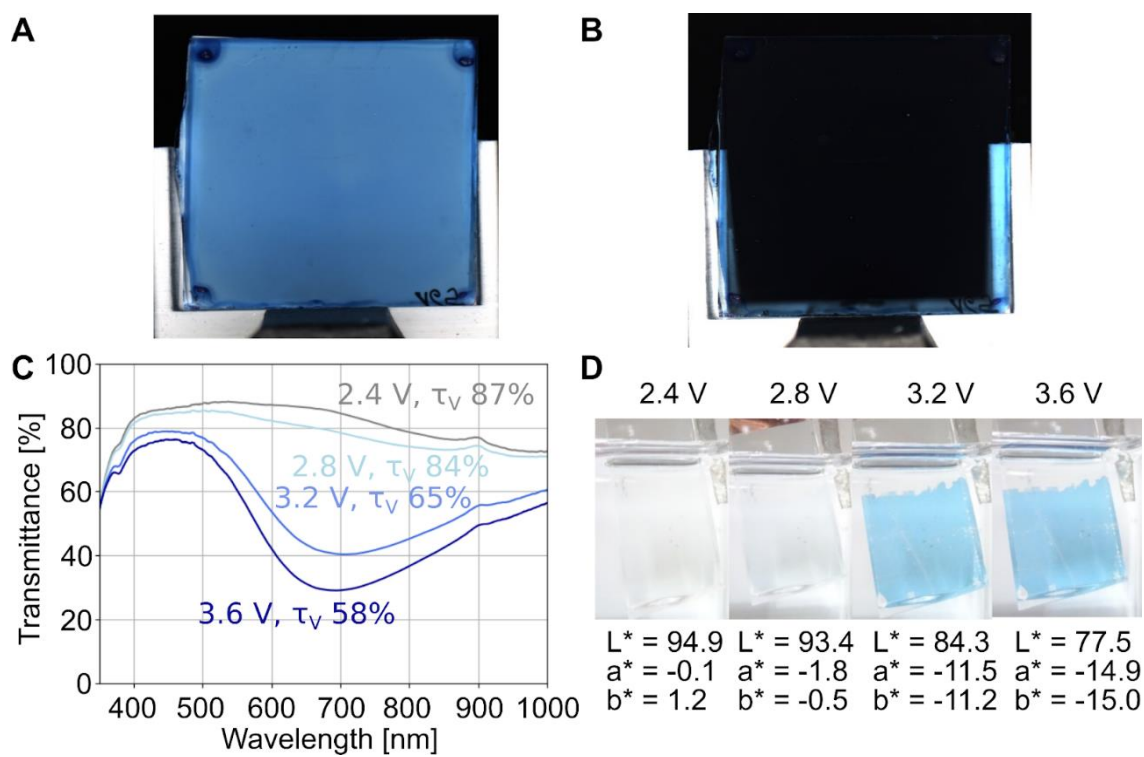

**Figure 6.** Photo of the PB thin film on FTO glass (optical density: 0.6, defect density: 0.13) against (A) a white background and (B) a black background. (C) Gradual transmittance changes upon application of a potential between 2.4 V (PW) and 3.6 V vs. Li/Li$^+$ (PB) with two intermediate states at 2.8 V and 3.2 V vs. Li/Li$^+$ in 1 M LiClO$_4$/PC. (D) Photos of the PB thin film at 2.4, 2.8, 3.2 and 3.6 V vs. Li/Li$^+$ and the respective L*a*b* color coordinates.

The photos of the two PB layers with a white background show the complete coating of the area, and the color intensity is represented by their optical densities (Figure 5A and Figure 6A). The absence of particles in the photos with a dark background demonstrates the low degree of defect density (Figure 5B and Figure 6B). The potential-dependent transmittance spectra of PB with higher optical density are shown in Figure 5C, and with lower optical density in Figure 6C. As an anodically-coloring EC material, PB switches reversibly from light blue (oxidized state, PB - Fe$^{2+}$/Fe$^{3+}$) to colorless (reduced state, Prussian white, PW - Fe$^{2+}$/Fe$^{2+}$) in the potential range from 2.4 V to 3.6 V vs. Li/Li$^+$. Further oxidation states of Prussian green (PG) and Prussian yellow (PY - Fe$^{3+/3+}$) can be reached at higher potentials, as demonstrated in other studies. [21, 22] Although both PB layers have different film thicknesses and thus different visible light transmittance $\tau_v$ in the colored state (38% vs. 58%), the bleached states of both samples reach $\tau_v = 87\%$, indicating a vastly colorless state at 2.4 V vs. Li/Li$^+$. The photos of the potential-dependent coloration states of the PB thin films are depicted in Figures 5D and 6D; they demonstrate the change of the color coordinates according to the $\tau_v$. In addition, two intermediate states at 2.8 V and 3.2 V vs. Li/Li$^+$ are measured to demonstrate the feature of EC thin films to reach various tint levels at specific potentials and voltages, respectively, allowing the adjustment of the transmittance and light control, e.g., for use in smart windows or other dimmable applications.

Cyclic voltammetry measurements of the PB layer with higher optical density (**Figure S7A**) and lower optical density (Figure S7B) are included in the SI, showing the anodic (oxidation to PB) and cathodic (reduction to PW) curves in the given potential range. Their comparison reveals higher peak currents and charge capacities due to the higher film thickness. The anodic peak of the sample with lower optical density shifts to more positive potentials, and the cathodic peak to more negative potentials. This results in an increased peak separation, indicating slower electron transfer kinetics.

### 3.5 On the versatility of the platform: Processing OLED thin films

The characterization methods and the concept of automated iterative parameter optimization can be adapted for other solution-processable thin film technologies. To demonstrate its versatility, we also used the platform to deposit and monitor functional layers that are often employed in OLEDs. Therefore, we deposited the hole transport layer poly(3,4-ethylenedioxythiophene):polystyrene sulfonate (PEDOT:PSS), followed by the emitter layer of the polymer Super Yellow (SY). All samples were spin coated atop the glass substrates with conductive indium tin oxide (ITO) layer and investigated for their layer quality using the methods described above.

Compared to ECDs, solution-processed OLEDs comprise rather thin and very homogeneous layers, with a thickness often well below 100 nm. For the detection and quantification of film defects in both the PEDOT:PSS layer and the subsequent Super Yellow layer, we used the same imaging setup without further modification. **Figure 7A** and Figure 7B show the photographs of PEDOT:PSS layer against a white and a black background. The darkfield image with a white background reveals macroscopic thickness variations (color gradients), but does not show individual defects or dust contamination, which become visible in the darkfield image with a black background. These defects can, in principle, damage the Super Yellow layer during subsequent deposition.

The other key optimization parameter of the light-emitting layers of OLEDs is their fluorescence. Both the spectral distribution and relative amplitude of the fluorescence reveal whether photoexcited charge carriers recombine efficiently radiatively (used for light emission in the device) or non-radiatively, which indicates losses. Here, we use changes in peak positions and the peak amplitudes of the fluorescence as an indirect metric for the layer thickness of the SY layer and for the layer conformation, such as polymer-chain alignment due to different drying speeds of the film. To determine the peak center wavelength, each sample was analysed by fitting three Gaussian curves (Figure 7E). This proof-of-concept analysis can provide feedback for a future closed-loop optimization of an OLED layer stack, but this would require switching to inert atmosphere conditions inside a glovebox in order to avoid the detrimental impact of water ingress.

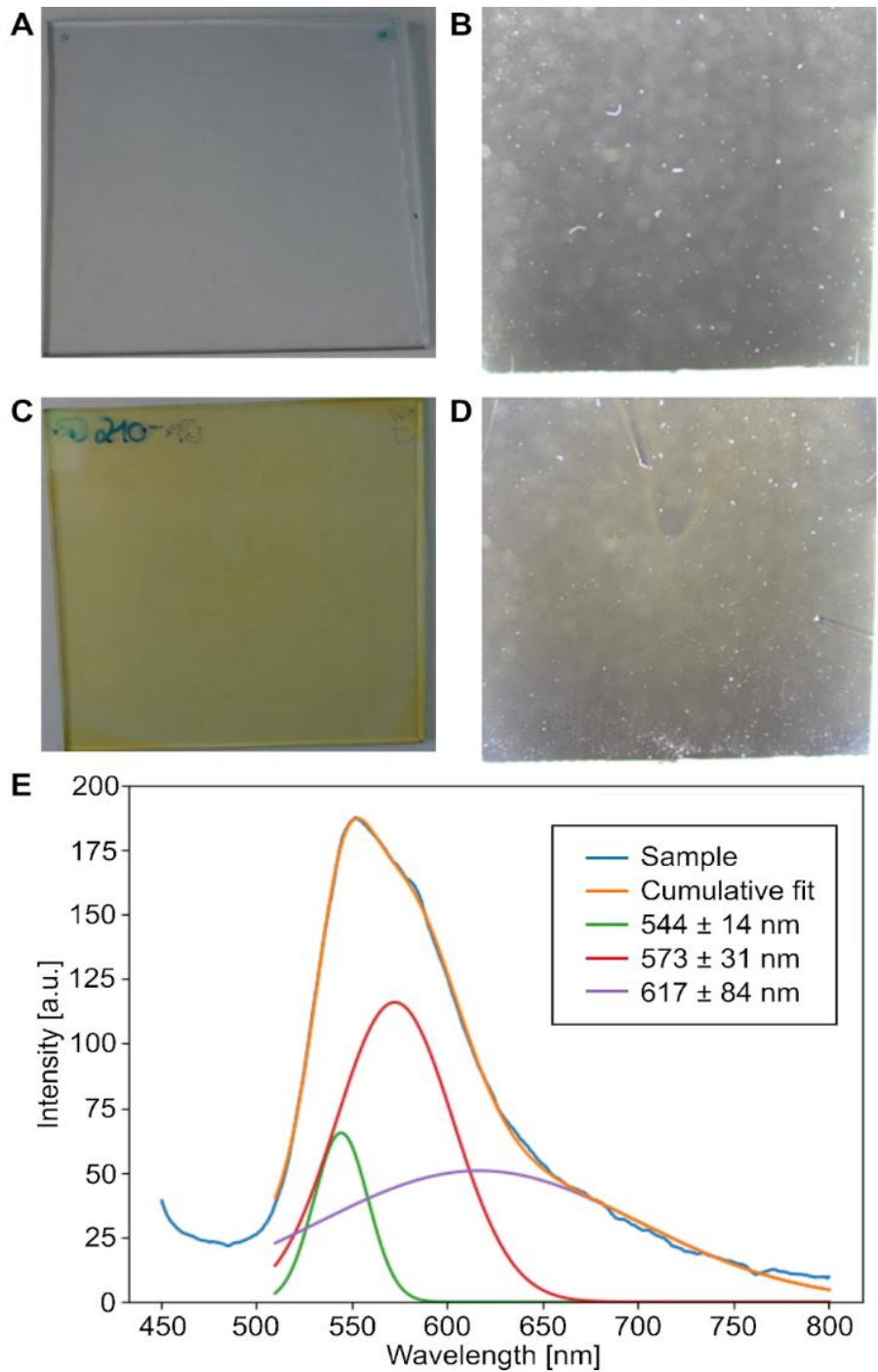

**Figure 7.** Photos of a PEDOT:PSS layer on an ITO-coated glass substrate (5x5 cm$^2$) against (A) a white and (B) a black background. The white background allows the visual differentiation of the film's color, while defects such as dust contamination in the sample become visible with black background images. C) SY atop the PEDOT:PSS layer coated at 1000 rpm for 30 s results in thickness inhomogeneities that become visible in color gradients in the photo with white background (D), while dust contamination and smaller defects are better visible in front of a black background; (E) Fluorescence spectra of SY layers spin coated atop PEDOT:PSS (1000 rpm for 30 s) and Gaussian peak analysis by decomposing the spectrum into three distinct peaks.

## 4. Conclusion and outlook

A modular and flexible automation platform for solution processing of thin films with machine learning-based decision-making was presented and used to optimize EC thin films. We demonstrated a systematic multi-objective optimization campaign with the objective of high optical density with low defect density. Compared to the initial data, we could increase the optical density six-fold, while keeping or even reducing the defect density. This was achieved through systematic optimization of processing conditions and systematic characterization to ensure uniformity, and at the same time minimize imperfections in the thin films. The automated approach streamlines the development, enhances reproducibility, and accelerates the optimization of material properties, ultimately leading to improved performance and efficiency in ECDs.

Specifically, we employed a machine learning-based decision-making algorithm based on Bayesian optimization to suggest a combination of spin speed, spin time, spin acceleration, and ink concentration. Future, more extensive optimizations will also encompass postprocessing conditions, e.g. annealing temperature and annealing time, while further studying the influence of ambient conditions such as humidity and temperature with more statistics. Furthermore, spatially resolved spectroscopic analysis, as well as electrical conductivity by four-point probe measurements, will be able to complement darkfield imaging for sample characterization and thus will add further optimization objectives for machine learning-based decision-making to create application-ready EC thin films. The applicability of our platform to other thin film materials furthermore opens opportunities for optimization of e.g. organic electronics materials.

## 5. Experimental Section/Methods

*Materials and substances*

Fluorine-doped tin oxide (FTO) glass from Pilkington (TEC$^{TM}$ 15, sheet resistance: approximately 15 Ω/sq) was used for the fabrication of the EC thin film electrodes. Prior to the coating, the 5x5 cm$^2$ glass sheets were cleaned with propan-2-ol, and corona treated with 0.15 kV to improve the adhesion of the aqueous PB dispersion. The corona treatment was performed no longer than 90 minutes before coating, depending on the sequence of samples in a batch of 20.

*Hardware - Modular solution processing platform*

The automated processing and characterization platform developed in this work (Figure 1) is structured modularly and incorporates a central robotic arm for sample transfer between stations. The stations include sample storage, ink storage, spin coating, annealing on a hot plate, and UV-vis characterization (Figure 1A). A darkfield imaging module was implemented in the platform, but was not fully integrated into the automated workflow. Therefore, measurements were performed manually. All stations are controlled by an industrial programmable logic controller (PLC, Siemens S7-1500) using multiple bus systems. To ensure user and machine safety, the PLC is connected to a PILZ system. Programming of the PLC was carried out using Siemens' Totally Integrated Automation Portal, where the production process was translated into a sequence of steps represented in function block diagrams. The system operates under clean room conditions (Figure 1B).

The POLOS SPS150i spin coater has been automated using reverse engineering. The spin coater's human-machine interface has been replaced by a microcontroller that can communicate with the PLC. The IKA C-MAG HS 7 control heating plate is controlled via RS-232, and the rLINE® 1-channel 5000 µl dispensing unit is also controlled via RS-232. The peristaltic pump, the pneumatic system for the spin coater, and the safety signals of the PILZ device are controlled by analog signals and edges. The GP7 robot arm from Yaskawa communicates with the PLC via Profinet using the Motologix protocol.

*Preparation of the PB dispersions*

The synthesis of the PB nanoparticles is described elsewhere.[23] The aqueous PB dispersions were used in concentrations of 2.4, 3.0, 3.4, and 4.0 wt.%, and Borchi Gen 1253 (aqueous solution of an acrylic ester copolymer) was added as a dispersing agent. The PB dispersion was kept stirring during storage. Before coating, the PB dispersions were filtered with a 0.8 µm syringe filter.

*PB thin film coating*

A coating volume of 3000 µL was dispensed on the conductive side of the FTO glass without rotating the spin table. Then, a specific spin acceleration was set to achieve the target spin speed and kept constant for a certain spin time. [19] After spin coating, the sample was transferred to the middle position of an aluminum block on the heating plate with a temperature of 230 °C to reach a constant temperature of approximately 100 °C on the surface of the PB-coated FTO glass. The PB thin film electrode was annealed for 1 minute and then cooled down in the storage rack to room temperature.

*PEDOT:PSS and SY thin film coatings*

The coating of the OLED included the PEDOT:PSS layer followed by the SY layer. The ITO-coated glass substrates were cleaned with propan-2-ol and dried with compressed air. The PEDOT:PSS dispersion was diluted at a ratio of 1:3 (vol:vol) in ethanol. Before spin coating, a volume of 400 µL was applied onto the ITO-electrode. The PEDOT:PSS layer was coated at a spin speed of 2000 rpm and a spin acceleration of 2000 rpm/s for 30 s. The samples were then annealed at 120 °C for 2 min. The SY (PDY-132, Merck) was coated atop the PEDOT:PSS layer from o-xylene solution (2 g/L). 200 µL of this solution were dispensed onto the substrate (ITO/PEDOT:PSS) and subsequently rotated 180° clockwise and counterclockwise for 10 s (acceleration: 2000 rpm/s) in order to evenly spread the solution. Immediately afterwards, the substrate was spun for 30 s at 2000 rpm and then dried for 10 s at a speed of 4000 rpm and an acceleration of 2000 rpm/s (SD210-10).

*Darkfield imaging*

All images were captured by a Blackfly BFS-U3-120S4C-CS camera equipped with an Edmund Optics 25 mm C Series Lens. The camera was positioned 7 cm above the substrate holder inside a dark casing, which was covered with black paper from Thorlabs. The C-mount lens was connected to the CS–mount camera using a Thorlabs C-Mount Extension adapter. An LED ring light SRL-12-WT-s (5000 K) from MBJ Imaging GmbH with an outside diameter of 121 mm and an inside diameter of 87 mm was used for illuminating the samples. The PB thin film was illuminated alternately against black and white backgrounds. The sample was illuminated from the direction of the camera at a 120° angle within a 360° plane to avoid any reflections on the sample. The focus of the lens was adjusted to the sample plane to illuminate particles against the black background at a distance of 8 cm for the images with a black background, with the lens opened to f/1.4. For the images, white photo paper was positioned beneath the sample. Images were taken using the SpinView software and analyzed automatically using classical computer vision methods (see Methods: Data Analysis).

*UV-vis spectrometry*

UV-vis measurements of the PB thin films were conducted using a CCD-based AvaSpec-3648 Fiber Optic Spectrometer (Avantes). The spectra were taken at the centre of the coated area, using a customized sample holder to ensure consistent positioning. The CIE L*a*b* color coordinates were calculated with the Avasoft 8 software and reproduced using a customizable Python code to be usable for optimization purposes, employing the standard illuminant D65 and the standard observer at 10°.

*Fluorescence spectroscopy*

The ITO/PEDOT:PSS/SY-coated glass substrates were examined by fluorescence spectroscopy. The SY was excited at a wavelength of 405 nm, and the detection was aligned to the excitation to maximize the fluorescence between 500 nm and 800 nm. We note that SY degraded under continuous excitation, which made an exchange of the samples necessary for repetitive measurements. Due to the strong swelling of the PEDOT:PSS layers by water and to avoid further contamination of the layers, the samples were stored in an inert atmosphere.

*Spectroelectrochemical and electrochemical characterization*

All measurements were performed with a Solartron Multistat 1470E potentiostat/galvanostat. Transmittance spectra were recorded with a CCD-based AvaSpec-3648 Fiber Optic Spectrometer (Avantes). The CIE L*a*b* color coordinates were calculated with the Avasoft 8 software with the standard illuminant D65 and the standard observer at 10°. The potential-dependent measurements of the PB layers on FTO glass (active area: 1 x 1 $cm^2$) were carried out with a three-electrode setup in a silica glass cuvette, using lithium metal as counter and reference electrodes, and 1 M lithium perchlorate in propylene carbonate ($LiClO_4$/PC) as electrolyte under argon atmosphere (glovebox) at room temperature. Electrical contact was provided by an adhesive copper tape. The potential was increased in 0.4 V steps from 2.4 V to 3.2 V vs. Li/$Li^+$, and the transmittance spectra at each potential were measured accordingly after no further change was observed. The visible light transmittance ($\tau_v$) was calculated according to DIN EN 410.

The cyclic voltammetry measurements of PB were performed after the spectroelectrochemical measurement in the same setup with a scan rate of 1 mV/s at 25 °C between 2.4 V and 3.6 V vs. Li/Li$^+$.

*Data analysis*

We used a traditional computer vision approach for defect analysis, which included several image processing steps. Initially, the images were cropped to isolate the samples' active area. They were then converted to greyscale and binarized using fixed thresholds to separate defective regions from non-defective ones. Defect density was assessed by calculating the area fractions of defective regions relative to the total active area. The whole code was implemented with the OpenCV library. Separate analyses were performed for images with bright and dark backgrounds, and the defect regions were employed as two distinct objectives in BO. For the sake of simplicity, in the figures of this manuscript, defect densities of dark and bright background images are summed up. Additional information about individual defect densities in bright and dark background images can be found in the SI. From the UV-vis spectroscopy data, we computed the optical density, defined as the maximum absorbance observed within the wavelength range of 650 nm to 900 nm. This was derived from the absorbance (A), calculated as -log10(T), where T (transmittance) was determined as (sample−dark)/(reference−dark). The code for automated image processing as well as UV-vis analysis can be found on GitHub (see code and data availability).

*Bayesian optimization*

To explore the Pareto front of low defect density and high optical density, we employed multi-objective BO. The optimization was performed over four hyperparameters: PB ink concentration, and three spin coater processing parameters—spin acceleration, spin speed, and spin time. A set of two independent Gaussian processes was used as surrogate models for the objective functions. As the acquisition function, we used qLogNEHVI, a numerically more stable variant of the well-established Noisy Expected Hypervolume Improvement (NEHVI).[24] In multi-objective optimization, the hypervolume serves as a scalar quality indicator, measuring the volume of the objective space dominated by a set of solutions—the Pareto front—relative to a predefined reference point. Building upon this, the expected hypervolume improvement quantifies how much this volume could increase by sampling a new, unexplored candidate solution. This guides the search towards promising regions in the objective space. Ten iterations were performed, with ten candidate parameter sets generated at each iteration. Each parameter set was evaluated in duplicate to quantify in-batch variation. In total, 100 distinct parameter sets were evaluated, resulting in 200 experimental samples processed across the campaign. All code was developed using the Botorch library.[25]

## Acknowledgements

We acknowledge support by the Federal Ministry of Education and Research (BMBF) under Grant No. 01DM21002 (FLAIM). The authors acknowledge support by the state of Baden-Württemberg through bwHPC.

## Conflict of Interest

The authors declare no conflict of interest.

## Author contributions

S.D. and L.T. contributed equally. All authors jointly developed the idea and prepared the manuscript. S.D., H.R., and L.T. conducted the experiments and analyzed the data. L.T. and P.F. were responsible for the software. P.F. and M.S. were responsible for the conceptualization, methodology, supervision, project administration, and funding acquisition.

## Data Availability Statement

The data collected in the campaign, as well as the code to reproduce the results of this paper, can be found on GitHub: https://github.com/aimat-lab/SDL_electrochromic.

# Supporting Information

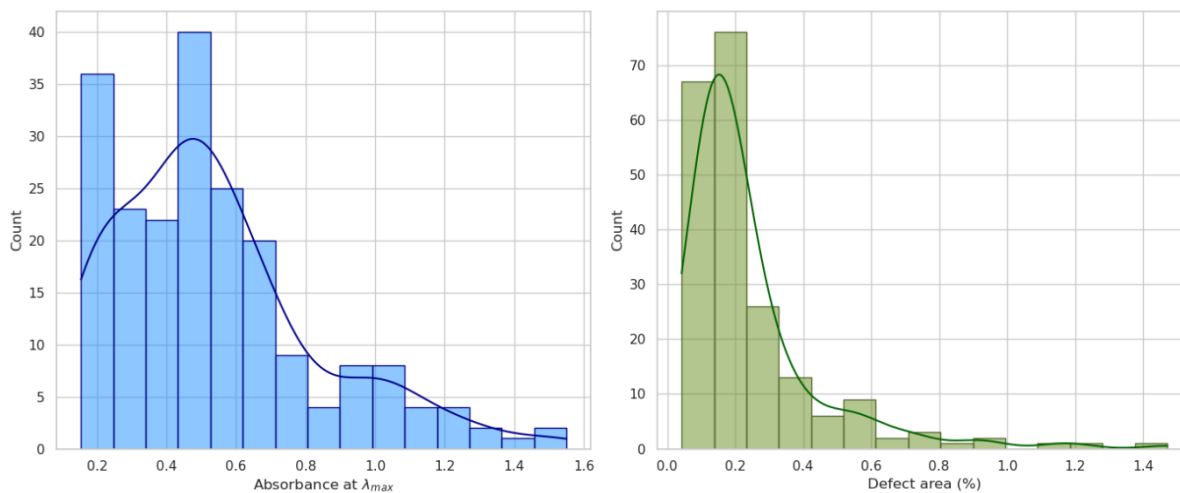

**Figure S1.** Distributions of the values of the two objective functions over all experiments in all iterations.

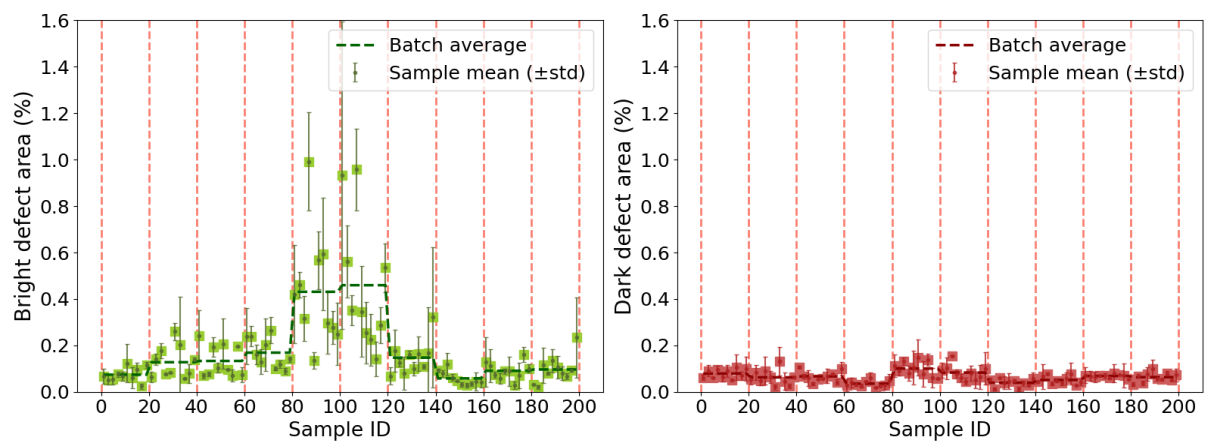

**Figure S2.** Contributions from bright (left) and dark (right) images to the relative defect area reported in Figure 3B.

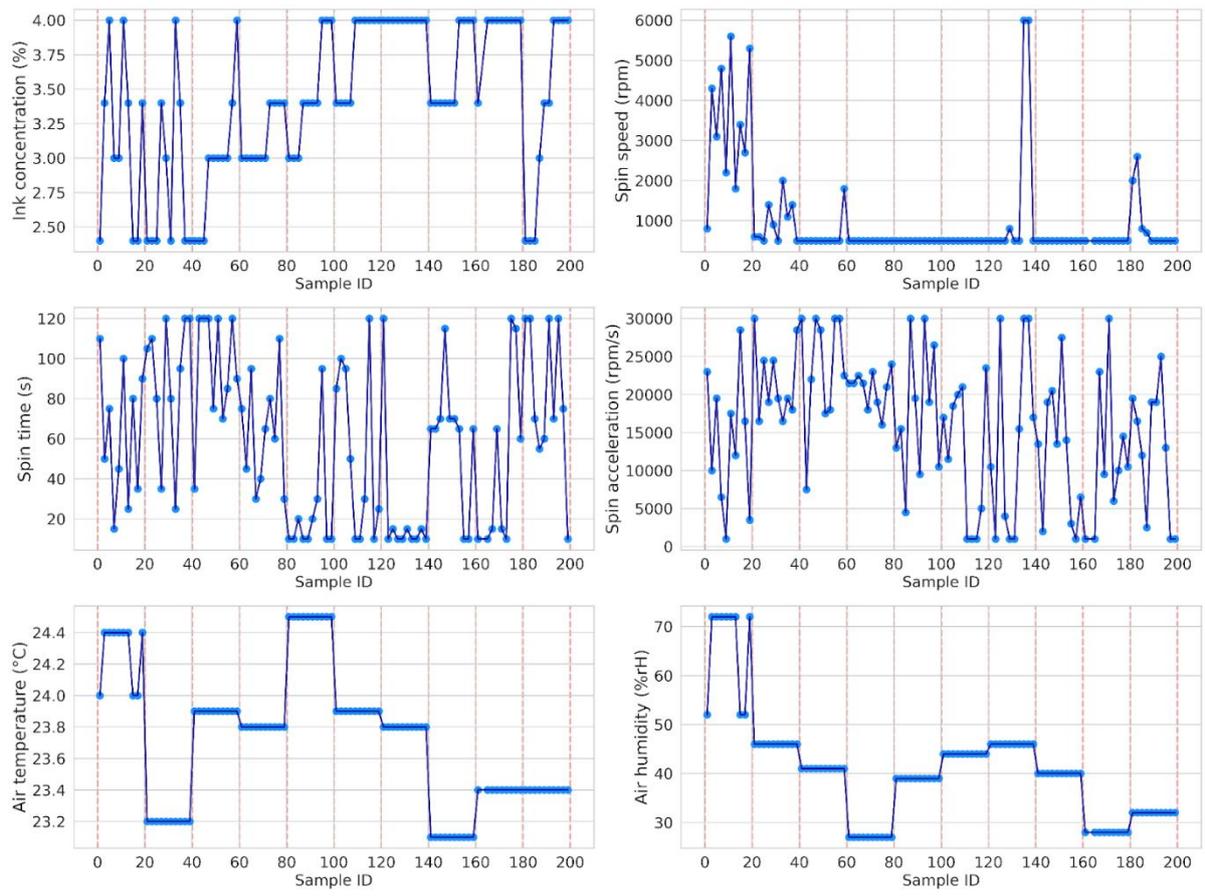

**Figure S3**. Processing parameters of each sample in the coating campaign.

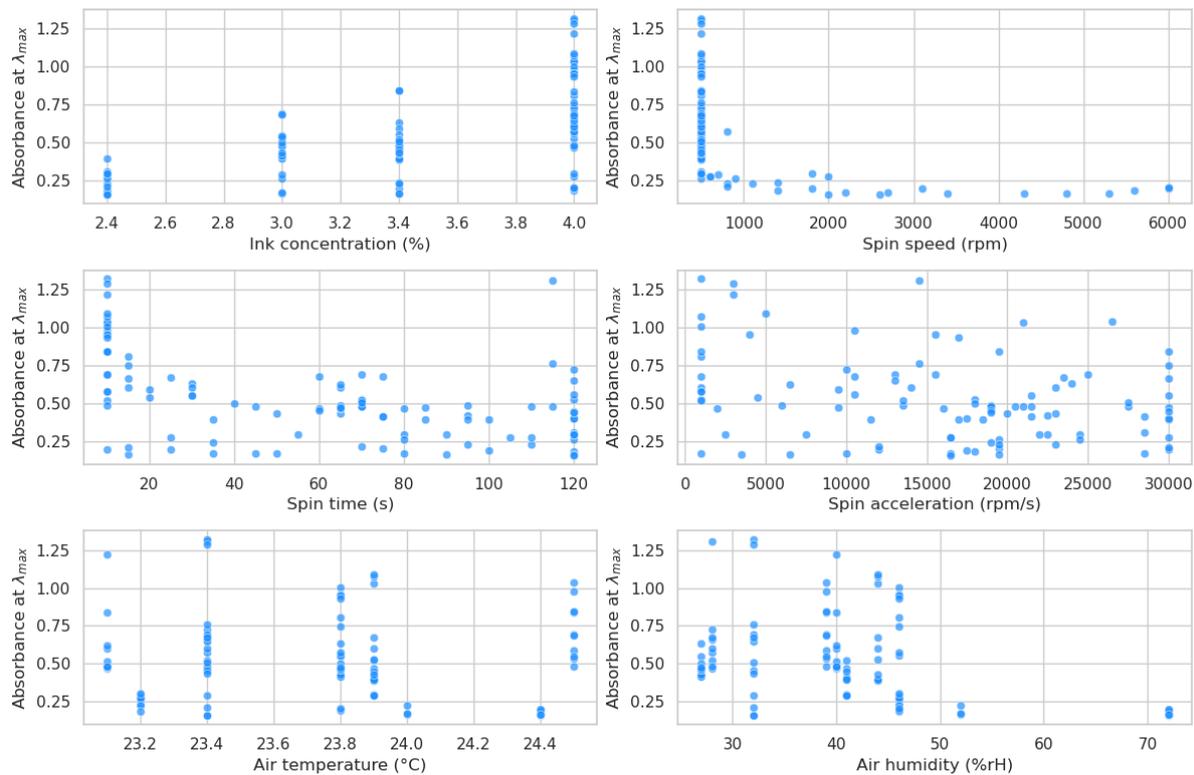

**Figure S4**. Optical density as a function of processing parameters.

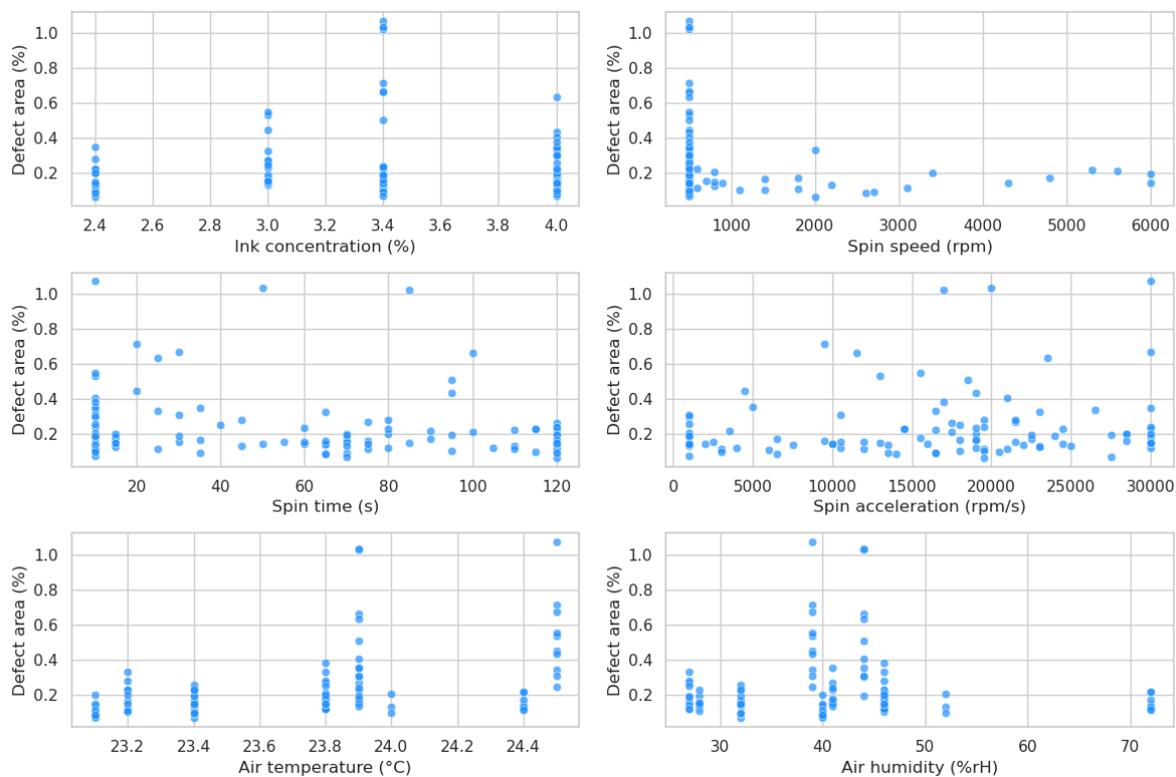

**Figure S5.** Defect area as a function of processing parameters.

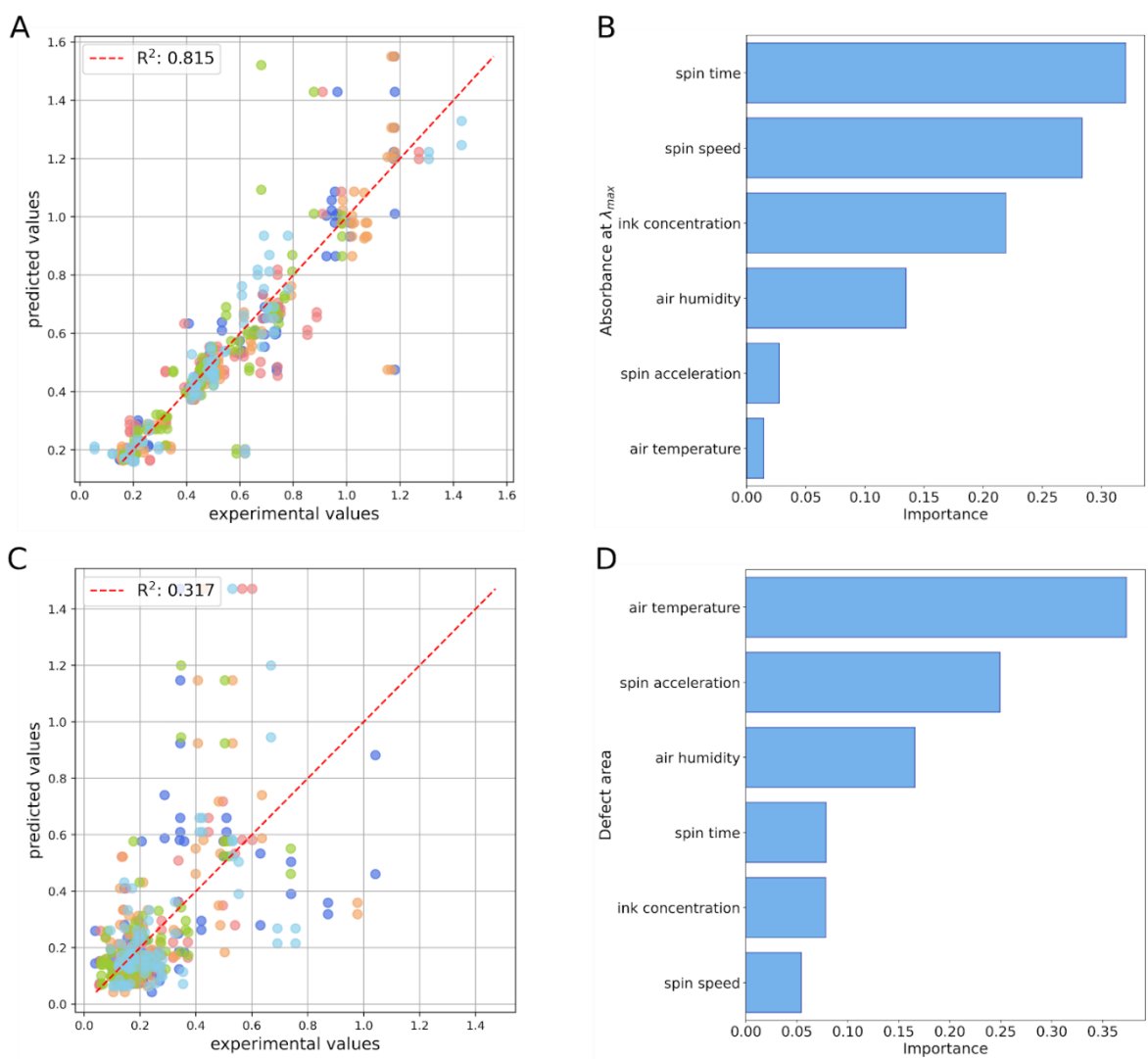

**Figure S6.** Feature importance and coefficient of determination on a cross-fold validation, including air temperature and humidity.

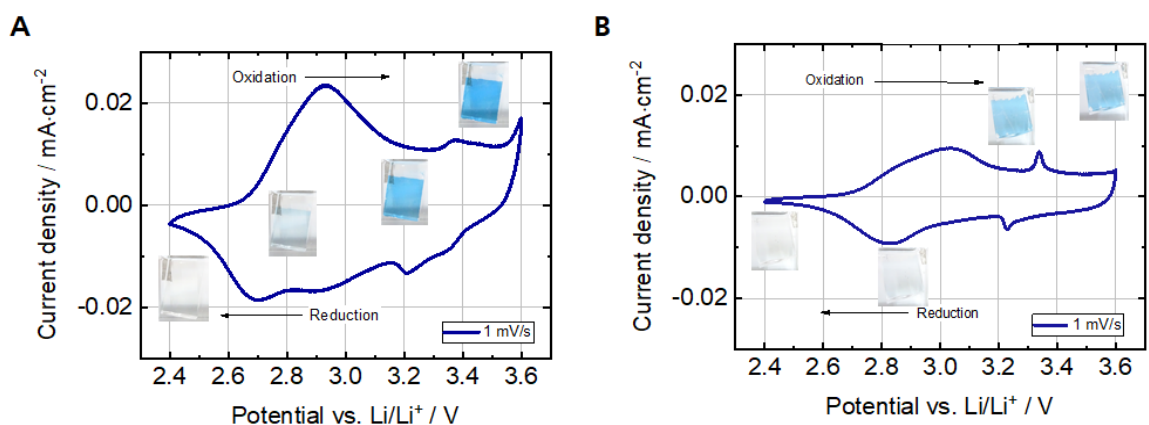

**Figure S7.** Cyclic voltammograms (scan rate: 1 mV/s) of PB on FTO glass as working electrode, Li as counter and reference electrode, and 1 M LiClO$_4$/PC as electrolyte. A) Optical density: 1.5, defect density: 0.15 and B) optical density: 0.6, defect density: 0.13.